\documentclass[letterpaper]{article}
\usepackage[preprint]{aaai2027}

\usepackage[hyphens]{url}
\usepackage{graphicx}
\usepackage{natbib}
\usepackage{caption}
\usepackage{booktabs}
\usepackage{amsmath}
\usepackage{amssymb}
\usepackage{array}
\usepackage{vcell}


\title{RayLift: Lifting Complementary Ray-Wise Evidence with 3D Geometry Priors
for Semantic Scene Completion}
\author{
  Meng Wang\equalcontrib,
  Hongxia Yu\equalcontrib,
  Wenzhe He,
  Xingdong Song,
  Huilong Pi,
  Jiapeng Zhang,
  Ruihui Li
}
\affiliations{
  College of Computer Science and Electronic Engineering,Hunan University,Changsha,China\\
  \{willem,yhx20252025,hewenzhe,songxingdong,phl880217,zhangjp,liruihui\}@hnu.edu.cn
}

\begin{document}

\maketitle
\begin{abstract}
Camera-based 3D semantic scene completion (SSC) provides comprehensive scene understanding for autonomous driving and robotics. However, existing methods often treat stereo depth estimates as deterministic geometric constraints, causing depth uncertainty and local correspondence errors to propagate directly into voxel representations. To address this issue, we propose RayLift, a framework that uses stereo geometry as a metric reference while incorporating complementary ray evidence to recover reliable 3D structures adaptively. RayLift first employs a Complementary Context Encoder that extracts geometry-aware priors from a frozen 3D vision foundation model, thereby enriching the scene context. It then introduces a Depth Ray Evidence Lifter module that jointly models geometric dissimilarity, depth confidence, and spatial uncertainty to adaptively sample and weight candidate surface locations along each camera ray. Finally, a Semantic-Aware Voxel Integrator injects the resulting ray evidence into voxel features by explicitly modeling their spatial support. Extensive experiments on SemanticKITTI and SSCBench-KITTI-360 demonstrate that RayLift achieves competitive performance and consistently outperforms existing methods.
\end{abstract}

\section{Introduction}

Semantic scene completion (SSC) reconstructs the geometry and semantic labels of an entire 3D scene from partial observations. This capability is essential for autonomous driving, robotics, and embodied intelligence, where an agent must reason about both visible and occluded space. Early SSC systems relied primarily on range sensors~\cite{Song_2017_CVPR,behley2019semantickitti}. Camera-based SSC has since emerged as a practical alternative because RGB cameras are inexpensive and widely deployed~\cite{cao2022monoscene,li2023sscbench}. However, images provide only two-dimensional appearance cues, so camera-based methods must establish reliable correspondences between image content and 3D voxels before performing scene-level reasoning.

\begin{figure}[t]
  \centering
  \includegraphics[width=0.95\columnwidth]{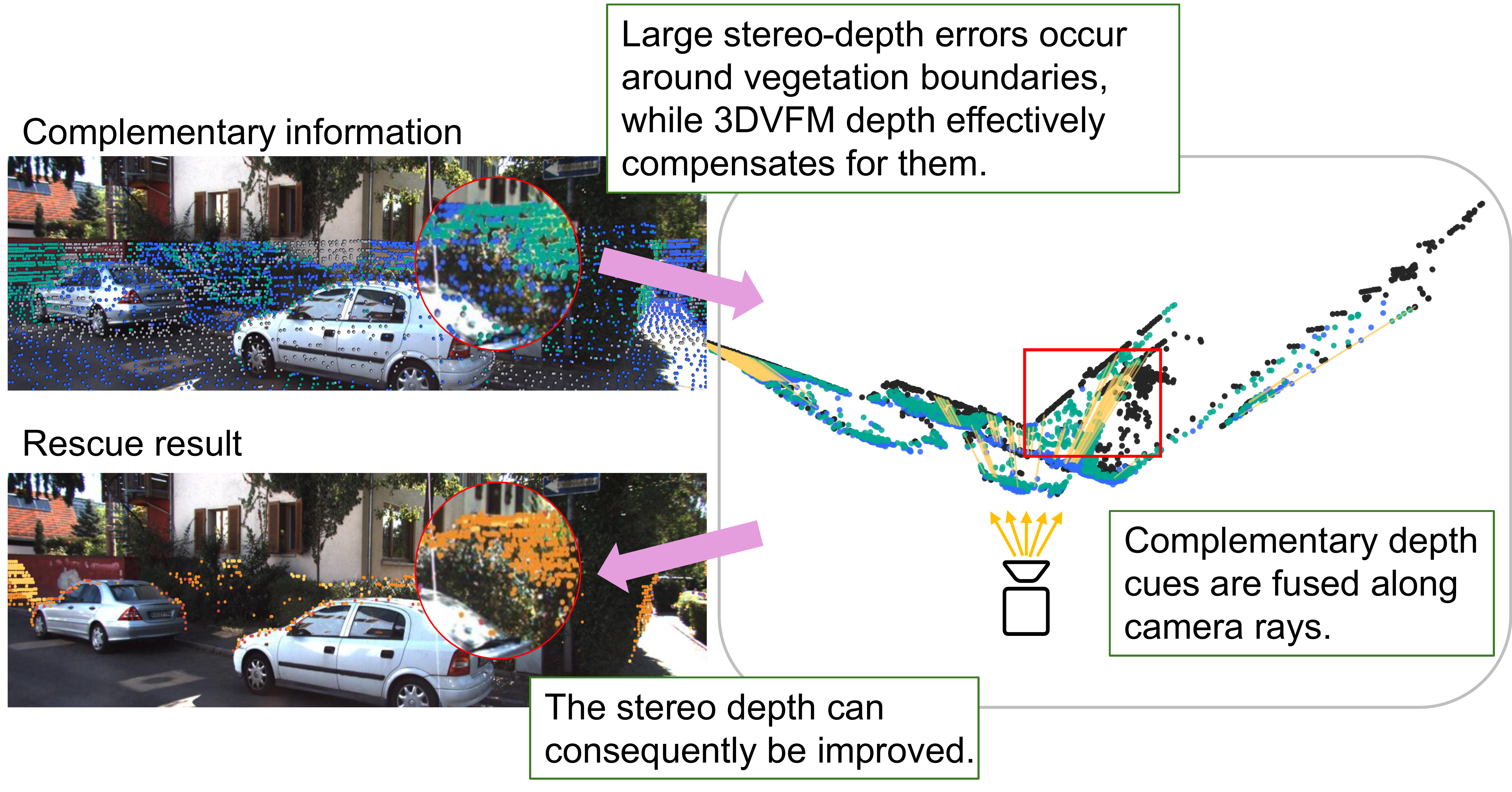}
  \caption{Motivation of RayLift. This figure illustrates how exploiting complementary depth cues can improve overall depth quality in an idealized setting. The green and blue points on the left indicate locations where 3DVFM depth and stereo depth are more accurate respectively. The yellow and orange points below represent stereo-depth errors corrected to different degrees. Our method is built upon this local complementarity between the two depth sources.}
  \label{fig:motivation}
\end{figure}

Existing camera-based SSC methods commonly extract image features, estimate depth, and lift the features into a 3D voxel representation~\cite{cao2022monoscene,philion2020lift,li2023sscbench}. Stereo systems strengthen this pipeline by providing metric depth from cross-view correspondence~\cite{li2023sscbench}. Most methods nevertheless reduce the stereo prediction to a single depth value and treat the resulting surface as a deterministic geometric constraint for feature lifting or voxel-query initialization. This design is effective when correspondence is reliable, but local errors remain common near object boundaries, weakly textured surfaces, and other ambiguous regions~\cite{scharstein2002taxonomy}. An erroneous depth can then place coherent image evidence at the wrong 3D locations and displace the reconstructed structure in the voxel grid.

Subsequent voxel networks can partially absorb such errors through contextual reasoning, but they rarely revisit the spatial location assigned during lifting. They also lack explicit alternative surface evidence when the stereo estimate is unreliable. Consequently, local correspondence errors propagate into the voxel representation before scene-level reasoning begins, where they are difficult to correct and can limit both geometric completion and semantic prediction.

Prior work addresses depth ambiguity mainly by modeling predictive uncertainty or by augmenting stereo features with additional geometric cues~\cite{li2023sscbench}. These cues typically reweight image features or support downstream reasoning, rather than forming localized 3D evidence that can challenge an unreliable stereo surface. Recent 3D vision foundation models offer a complementary opportunity~\cite{wang2025vggt}. Their pretrained representations encode transferable scene structure, while their monocular depth predictions provide an alternative geometric hypothesis. The central question is therefore how to convert these heterogeneous priors into spatially localized evidence without discarding the metric reference supplied by stereo geometry.

Motivated by this question, we propose RayLift, a camera-based SSC framework that retains stereo geometry as a metric reference while converting complementary visual geometry priors into ray-wise voxel evidence. RayLift operates at three levels. First, the Complementary Context Encoder adapts multi-level geometry-aware features from a frozen 3D vision foundation model and combines them with task-specific image features. Second, the Depth Ray Evidence Lifter aligns the stereo surface and the foundation-model surface along each camera ray. It models their geometric discrepancy, depth confidence, and spatial uncertainty to sample candidate locations and assign evidence weights adaptively. Third, the Semantic-Aware Voxel Integrator injects the resulting evidence into the voxel representation according to its spatial support. This design preserves reliable stereo geometry while providing localized alternatives where deterministic lifting is uncertain.

We evaluate RayLift on SemanticKITTI and SSCBench-KITTI-360 under matched experimental settings. The consistent improvements in occupancy IoU and semantic mIoU indicate that localized ray evidence benefits geometric completion without sacrificing semantic discrimination. Component analyses further isolate the roles of geometry-aware context, ray-wise evidence construction, and voxel integration.

Our main contributions are summarized as follows:
\begin{itemize}
  \item We introduce RayLift, a camera-based SSC framework that treats stereo depth as a metric reference rather than a deterministic constraint and converts complementary geometry priors into localized ray-wise voxel evidence.
  \item We design a Depth Ray Evidence Lifter that jointly models geometric discrepancy, depth confidence, and spatial uncertainty to sample candidate surface locations and assign their evidence weights adaptively.
  \item We develop a Complementary Context Encoder and a Semantic-Aware Voxel Integrator that respectively construct geometry-aware scene context and inject ray evidence into voxels according to its spatial support.
  \item Extensive experiments on SemanticKITTI and SSCBench-KITTI-360 demonstrate that RayLift achieves competitive performance and consistently outperforms existing methods.
\end{itemize}

\section{Related Work}
\subsection{3D Vision Foundation Models}

3D vision foundation models learned transferable geometric representations from large-scale visual data. Classical reconstruction pipelines relied heavily on multi-view geometry and iterative optimization~\cite{Schonberger_2016_CVPR}, whereas recent feed-forward models directly inferred scene geometry and camera parameters from images. DUSt3R~\cite{wang2024dust3r} and MASt3R~\cite{leroy2024grounding} were representative approaches that inferred coupled scene representations from image pairs and demonstrated the potential of end-to-end learned reconstruction. VGGT~\cite{wang2025vggt} extended this paradigm to joint multi-view processing and produced transferable geometry-aware features.

This line of research expanded toward increasingly general geometric perception. The Depth Anything family learned robust monocular depth representations at scale~\cite{lin2025depth,yang2024depth}. Subsequent variants explored metric depth estimation, camera pose estimation, and feed-forward 3D reconstruction. MapAnything~\cite{keetha2026mapanything} focused on constructing general spatial maps from visual observations and illustrated the potential of foundation models for map-level geometric reasoning. MoGe-2~\cite{wang2026moge} predicted dense 3D geometric cues from monocular images and provided a new representation for downstream metric and structure-aware scene understanding. More recently, VGGT-Omega~\cite{wang2026vggt} further increased model capacity and training diversity, strengthening the transferability of VGGT-style geometric representations to downstream 3D perception tasks.

\subsection{Camera-Based 3D Semantic Scene Completion}

Camera-based SSC constructed dense 3D voxel representations from RGB images without direct 3D measurements. MonoScene~\cite{cao2022monoscene} established an early camera-only framework by projecting image features into voxel space. Subsequent methods developed more effective 3D representations. TPVFormer~\cite{huang2023tri} used tri-perspective features to reduce dense voxel computation, while VoxFormer~\cite{li2023voxformer} followed a sparse-to-dense prediction pipeline. OccFormer~\cite{zhang2023occformer} combined lift-splat-shoot with Transformer-based 3D reasoning. Together, these methods integrated image-feature lifting, 3D spatial representation, and geometry--semantic reasoning into unified SSC pipelines.

Later studies improved this pipeline from several complementary directions. BRGScene~\cite{li2023bridging}, DepthSSC~\cite{yao2025depthssc}, and CGFormer~\cite{yu2024context} strengthened the alignment and interaction among image features, depth geometry, and voxel representations. HASSC~\cite{wang2024not} focused learning on ambiguous voxels, while IAMSSC~\cite{xiao2024instance} and Symphonies~\cite{jiang2023symphonize} introduced instance-level features or queries to facilitate 2D--3D interaction. VLScene~\cite{wang2025vlscene} transferred knowledge from vision--language models, whereas SGFormer~\cite{guo2025sgformer} incorporated satellite observations when available. Ocean~\cite{wang2026objectcentric} further improved instance-aware geometric and semantic reasoning. VoxDet~\cite{li2026voxdet} reformulated SSC as dense 3D detection to explicitly model instance locations and semantic categories during voxel prediction.

\begin{figure*}[t]
  \centering
  \includegraphics[width=\textwidth]{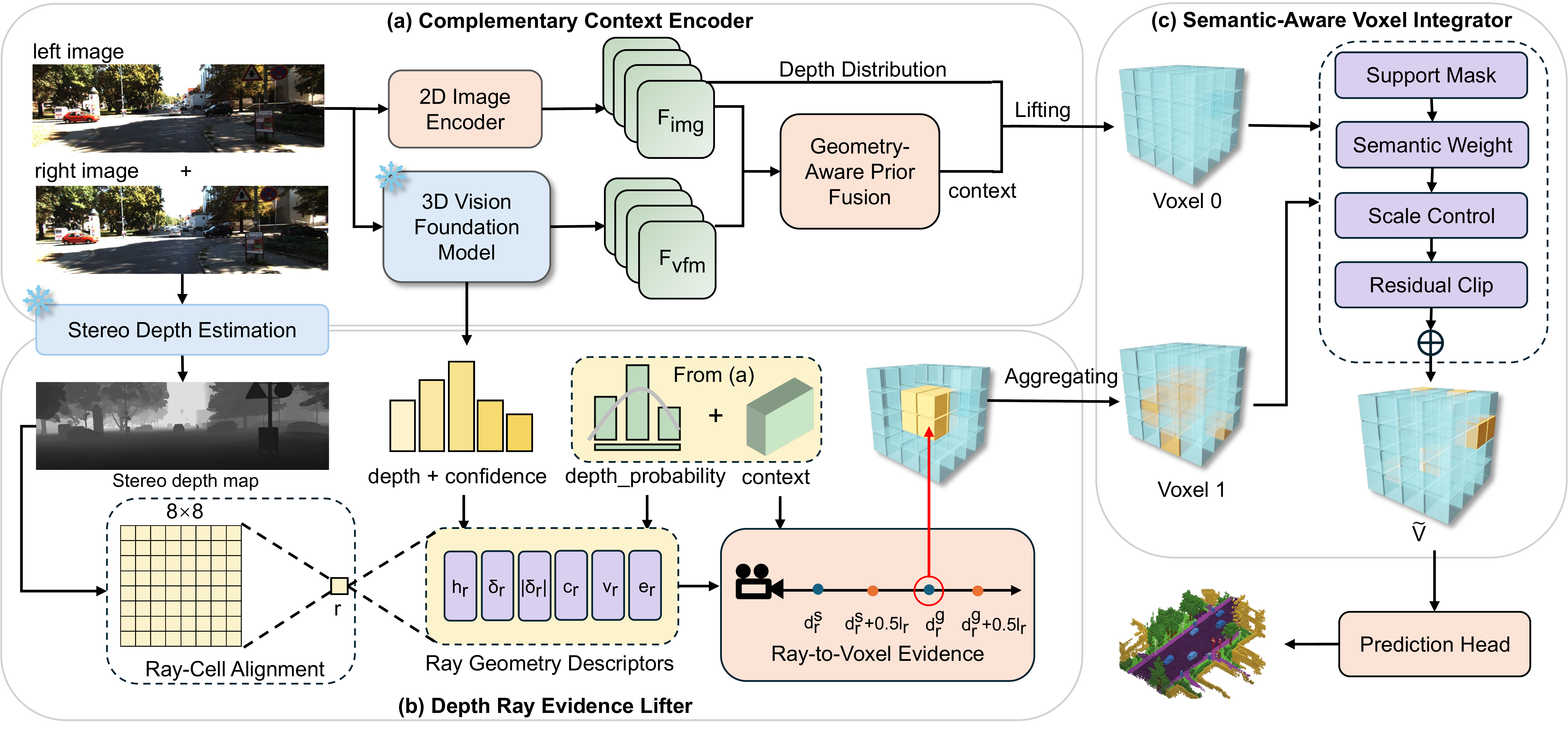}
  \caption{Overall architecture of RayLift. (a) The Complementary Context Encoder extracts multi-level geometry-aware features and fuses them with task-specific image features to construct a scene-context representation. (b) The Depth Ray Evidence Lifter aligns the surfaces estimated by stereo vision and the 3D geometry model within ray-grid cells, encodes their uncertainty and geometric relationship, samples weighted feature points, and maps them into ray-wise voxel evidence. (c) The Semantic-Aware Voxel Integrator fuses the ray evidence with the voxel representation and forwards the result to the prediction head.}
  \label{fig:framework}
\end{figure*}

\section{Methodology}

Given a stereo image pair, the 2D image encoder produces image features $F_{\mathrm{img}}$, while the stereo depth estimator provides metric depth $D_s$. The visual geometry model additionally supplies multi-level geometry-aware features $F_{\mathrm{vfm}}$, a depth prediction, and its confidence. The Complementary Context Encoder (CoCE) adapts and fuses $F_{\mathrm{img}}$ and $F_{\mathrm{vfm}}$ to construct the geometry-enhanced image context $C$. Following the existing depth-modeling pathway, $C$ is lifted according to the depth probability distribution $P_d$ and processed to obtain the initial scene voxel representation $V_0$. The Depth Ray Evidence Lifter (DREL) organizes $C$, $P_d$, $D_s$, and the visual-geometry depth and confidence along camera rays. It then samples contextual features and projects them into 3D space to construct complementary ray-wise voxel evidence $V_1$.  Finally, the Semantic-Aware Voxel Integrator (SAVI) integrates $V_1$ with $V_0$ before passing the unified representation to the prediction head.
Figure~\ref{fig:framework} illustrates the overall architecture of RayLift. 

\subsection{Complementary Context Encoder}

The 2D-to-3D lifting process assigns image context to 3D locations according to the predicted depth distribution. Because the subsequent ray-wise evidence construction evaluates candidate surface locations along camera rays, we complement the task-specific image context with structural cues learned through reconstruction pretraining. We employ the pretrained VGGT-$\Omega$~\cite{wang2026vggt} model to provide multi-level geometry-aware features, dense depth predictions, and confidence estimates for the subsequent stages.

We consider two feature sources within VGGT-$\Omega$: the DINOv3 backbone and the geometry aggregator~\cite{simeoni2025dinov3}. We evaluate both variants using the same architecture and optimization settings to isolate the effect of the feature source. The intermediate features of the geometry aggregator contain richer geometry-aware information and are better suited to RayLift than the general-purpose visual features produced by the backbone. We therefore extract features from multiple depths of the geometry aggregator, retaining local structure from shallower layers and scene-level relationships from deeper layers. The selected patch features are projected into a common feature space, spatially aligned with the 2D image features, and aggregated along the channel dimension to form the multi-level geometric representation.

\paragraph{Geometry-Aware Prior Fusion.}
The subsequent ray-evidence construction samples contextual features at candidate surface locations. These samples should preserve the task-specific semantic information in $F_{\mathrm{img}}$ while incorporating the structural cues associated with the geometry predictions in $F_{\mathrm{vfm}}$. We therefore fuse the two representations before lifting them into 3D space. After spatial and channel alignment, a lightweight modulation network jointly processes the two features and predicts a position- and channel-dependent weight $\omega$. This adaptive weighting controls how much geometry-aware information is introduced at each feature location and channel. The modulated geometric feature is injected into the image context as a residual increment scaled by a fixed factor $\alpha$. The resulting context is
\begin{equation}
C=H_c\!\left(
\rho(F_{\mathrm{img}})
+\alpha\omega\odot F_{\mathrm{vfm}};\eta
\right),
\label{eq:coce}
\end{equation}
where $\rho(\cdot)$ transforms the 2D image feature, $H_c$ denotes the context-generation head modulated by camera parameters $\eta$, and $\odot$ represents element-wise multiplication. The context $C$ is subsequently lifted into 3D space according to the depth probability distribution using a standard 2D-to-3D view transformation. The lifted features are further processed through voxel aggregation and image--voxel interaction to obtain the initial scene voxel representation $V_0$.

\subsection{Depth Ray Evidence Lifter}

We construct $V_0$ using a standard 2D-to-3D lifting scheme~\cite{philion2020lift}. The image context is projected into a 3D frustum according to the depth probability distribution, and voxel pooling subsequently maps the frustum features onto a 3D voxel grid. This pathway represents depth uncertainty through the probability distribution along each camera ray. However, the stereo and visual-geometry depths define two distinct candidate surfaces. Incorporating them only through a shared depth distribution does not explicitly preserve their relative displacement or individual reliability. Moreover, depth regression or weighted averaging may place the resulting surface estimate between the two candidates, weakening their geometric distinction.

To retain the relationship between the stereo and visual-geometry surfaces, we construct an independent ray-evidence pathway. Camera rays provide a common geometric coordinate along which the two depth estimates can be aligned and compared. Their relative positions characterize geometric disagreement, while the associated confidence indicates the reliability of the visual-geometry estimate. DREL uses these cues to sample locations near the candidate surfaces. Each sample is assigned a context feature produced by CoCE and a write weight that determines its contribution to the voxel representation. The samples are then back-projected and aggregated into localized ray-wise voxel evidence $V_1$, which complements the initial representation $V_0$.

Before constructing the ray cells, we calibrate the visual-geometry depth to the metric scale. The input depth maps have a resolution of $H\times W$, whereas the context and depth probability distribution produced by CoCE have a resolution of $H/8\times W/8$. We therefore define each $8\times8$ image region as a ray cell corresponding to one spatial location in the context and depth probability maps. We also evaluate a denser ray-cell partition. Although smaller cells increase the spatial coverage of the resulting 3D evidence, they do not improve prediction accuracy, as shown in the ablation study.

Each ray cell requires representative stereo depth, visual-geometry depth, and confidence values. We select the pixel with the minimum valid stereo depth and sample the visual-geometry depth and confidence at the same pixel location. This operation preserves the frontmost visible surface when a cell spans multiple depth layers or a depth discontinuity. It also provides a consistent image location at which the two candidate surfaces can be compared.

\paragraph{Ray Geometry Descriptor.}
We represent each ray cell $r$ using the geometric descriptor
\begin{equation}
q_r=
\left[
h_r,\,
\delta_r,\,
|\delta_r|,\,
c_r,\,
v_r,\,
e_r
\right].
\label{eq:ray_descriptor}
\end{equation}
Here, $h_r$ denotes the normalized entropy of the depth probability distribution $P_r(d)$. A small $h_r$ indicates that the probability mass is concentrated within a narrow depth interval, whereas a large value indicates a more dispersed distribution along the ray. The signed log-depth discrepancy $\delta_r$ encodes the relative front-to-back ordering of the visual-geometry and stereo surfaces. Its logarithmic form expresses the discrepancy relative to scene depth, reducing the dependence of its magnitude on the absolute range. The corresponding absolute value $|\delta_r|$ measures the separation between the two candidate surfaces independently of direction.

The relative confidence $c_r$ is obtained by log-transforming the raw visual-geometry confidence and normalizing it over the valid ray cells. It therefore indicates the reliability of the visual-geometry surface relative to other candidate surfaces. The binary indicator $v_r$ records whether the candidate depths are valid, preventing an invalid candidate pair from being interpreted as two geometrically consistent surfaces.

The final component $e_r$ measures the relative variation between the maximum and minimum valid stereo depths within the current $8\times8$ cell:
\begin{equation}
e_r=
\frac{d_r^{\max}-d_r^{\min}}
{d_r^{\min}+\epsilon},
\label{eq:local_depth_variation}
\end{equation}
where $d_r^{\min}$ and $d_r^{\max}$ denote the minimum and maximum finite positive stereo depths within the cell, $\epsilon$ is a small constant for numerical stability. A small $e_r$ indicates limited depth variation around the representative surface. In contrast, a large value suggests that the cell may contain foreground and background surfaces, an object or occlusion boundary, or a thin structure. The descriptor $q_r$ therefore allows the network to distinguish locally stable regions from depth discontinuities and adjust the sampling range and evidence weights accordingly.See supplementary materials for mathematical definitions and details of all parameters.

\paragraph{Ray-to-Voxel Evidence Construction.}
The stereo depth and scale-corrected visual-geometry depth define two candidate surfaces along each ray. Sampling only at these fixed surfaces provides limited tolerance to local depth errors and voxel discretization. Conversely, increasing the number of samples indiscriminately introduces redundant points and may distribute evidence far from both candidate surfaces. We therefore retain one sample at each surface and place one additional sample behind it, yielding $K=4$ samples per ray:
\begin{equation}
\mathcal D_r=
\left\{
d_r^s,\,
d_r^s+0.5\ell_r,\,
d_r^g,\,
d_r^g+0.5\ell_r
\right\},
\label{eq:ray_samples}
\end{equation}

where $d_r^s$ and $d_r^g$ denote the stereo and scale-corrected visual-geometry depths, respectively. The range $\ell_r$ remains ray-adaptive and is learned end-to-end.

To construct a voxel representation $V_1$ that can be integrated with $V_0$, each sample encodes its 3D position, feature content, and write strength. First, the sampled depth is back-projected using its image location and the camera parameters, producing the 3D position $x_{r,k}$. Second, a lightweight feature encoder maps the corresponding context feature to the ray feature $f_{r,k}$. A learnable sample-position embedding is added to distinguish the two candidate surfaces and their neighboring samples. Finally, the initial write strength predicted from the context and ray descriptor is adjusted according to depth validity and visual-geometry confidence. This operation produces the final weight $\widehat{\omega}_{r,k}$. The $k$-th sample on ray $r$ is therefore represented as
\begin{equation}
s_{r,k}=(x_{r,k},f_{r,k},\widehat{\omega}_{r,k}),
\quad d_{r,k}\in\mathcal D_r,\;
k=1,\ldots,K.
\label{eq:ray_sample}
\end{equation}
We distribute each sample feature to its neighboring voxels through trilinear projection. Contributions to the same voxel are aggregated according to $\widehat{\omega}_{r,k}$. A lightweight 3D convolutional encoder then organizes the discrete samples within their local voxel neighborhoods, producing the ray-wise voxel evidence $V_1$ and its validity mask $M$. Both are passed to SAVI, which integrates $V_1$ with the initial voxel representation $V_0$.

\newcommand{\semantickittitable}{%
\begin{table*}[!t]
\centering
\small
\setlength{\tabcolsep}{2pt}

\begin{tabular}{@{}lcc*{19}{c}@{}}
\toprule
Method & IoU & mIoU &
\rotatebox{90}{road} & \rotatebox{90}{sidewalk} &
\rotatebox{90}{parking} & \rotatebox{90}{other-grnd.} &
\rotatebox{90}{building} & \rotatebox{90}{car} &
\rotatebox{90}{truck} & \rotatebox{90}{bicycle} &
\rotatebox{90}{motorcycle} & \rotatebox{90}{other-veh.} &
\rotatebox{90}{vegetation} & \rotatebox{90}{trunk} &
\rotatebox{90}{terrain} & \rotatebox{90}{person} &
\rotatebox{90}{bicyclist} & \rotatebox{90}{motorcyclist} &
\rotatebox{90}{fence} & \rotatebox{90}{pole} &
\rotatebox{90}{traf.-sign} \\
\midrule
MonoScene & 34.16 & 11.08 & 54.7&27.1&24.8&5.7&14.4&18.8&3.3&0.5&0.7&4.4&14.9&2.4&19.5&1.0&1.4&0.4&11.1&3.3&2.1\\
VoxFormer & 43.21 & 13.41 & 54.1&26.9&25.1&7.3&23.5&21.7&3.6&1.9&1.6&4.1&24.4&8.1&24.2&1.6&1.1&0.0&13.1&6.6&5.7\\
OccFormer & 34.53 & 12.32 & 55.9&30.3&31.5&6.5&15.7&21.6&1.2&1.5&1.7&3.2&16.8&3.9&21.3&2.2&1.1&0.2&11.9&3.8&3.7\\
Symphonies & 42.19 & 15.04 & 58.4&29.3&26.9&11.7&24.7&23.6&3.2&3.6&2.6&5.6&24.2&10.0&23.1&\textbf{3.2}&1.9&\textbf{2.0}&16.1&7.7&8.0\\
CGFormer & 44.41 & 16.63 & 64.3&34.2&34.1&12.1&25.8&26.1&4.3&3.7&1.3&2.7&24.5&11.2&29.3&1.7&3.6&0.4&18.7&8.7&9.3\\
VLScene & 45.14 & 17.52 & 64.7&34.7&32.4&13.1&27.3&26.1&6.5&4.2&3.8&\textbf{8.3}&26.4&10.0&29.4&2.8&5.1&0.9&20.0&8.9&8.4\\
Ocean&45.62&17.40&65.1&34.9&33.7&12.8&25.4&26.8&5.2&4.9&1.9&5.6&26.9&11.6&30.7&2.3&2.2&1.7&20.8&8.7&9.5\\

\midrule
VoxDet\textsuperscript{$\dagger$} & 47.28&18.00&64.0&35.3&\textbf{35.1}&12.9&28.8&26.5&5.2&5.1&3.9&4.2&29.6&12.4&32.0&2.9&3.6&0.6&22.0&8.5&9.6\\
RayLift (VGGT-$\Omega$) & \textbf{48.38}&18.54&65.3&35.7&33.0&13.0&28.7&\textbf{27.1}&6.2&\textbf{5.7}&\textbf{5.5}&4.2&\textbf{30.8}&\textbf{12.8}&\textbf{32.1}&3.0&\textbf{6.5}&1.3&22.9&8.8&9.7\\
RayLift (MoGe-2) & 48.00&\textbf{18.71}&\textbf{65.5}&\textbf{35.9}&34.7&\textbf{14.7}&\textbf{29.0}&\textbf{27.1}&\textbf{7.2}&\textbf{5.7}&4.4&6.7&30.2&12.7&31.4&3.0&3.5&1.0&\textbf{23.0}&\textbf{9.3}&\textbf{10.5}\\
\bottomrule
\end{tabular}
\caption{Quantitative results on SemanticKITTI hidden test set. Best values in \textbf{bold}. \(\dagger\) denotes results reproduced using the official code implementation.}
\label{tab:semantickitti_main}
\end{table*}
}

\newcommand{\kittiresulttable}{%
\begin{table*}[!t]
\centering
\small
\setlength{\tabcolsep}{2pt}
\begin{tabular}{@{}lcc*{18}{c}@{}}
\toprule
Method & IoU & mIoU &
\rotatebox{90}{car} & \rotatebox{90}{bicycle} &
\rotatebox{90}{motorcycle} & \rotatebox{90}{truck} &
\rotatebox{90}{other-veh.} & \rotatebox{90}{person} &
\rotatebox{90}{road} & \rotatebox{90}{parking} &
\rotatebox{90}{sidewalk} & \rotatebox{90}{other-grnd.} &
\rotatebox{90}{building} & \rotatebox{90}{fence} &
\rotatebox{90}{vegetation} & \rotatebox{90}{terrain} &
\rotatebox{90}{pole} & \rotatebox{90}{traf.-sign} &
\rotatebox{90}{other-struct.} & \rotatebox{90}{other-obj.} \\
\midrule
MonoScene & 37.87&12.31&19.3&0.4&0.6&8.0&2.0&0.9&48.4&11.4&28.1&3.3&32.9&3.5&26.2&16.8&6.9&5.7&4.2&3.1\\
VoxFormer & 38.76&11.91&17.8&1.2&0.9&4.6&2.1&1.6&47.0&9.7&27.2&2.9&31.2&5.0&29.0&14.7&6.5&6.9&3.8&2.4\\
OccFormer & 40.27&13.81&22.6&0.7&0.3&9.9&3.8&2.8&54.3&13.4&31.5&3.6&36.4&4.8&31.0&19.5&7.8&8.5&7.0&4.6\\
Symphonies & 44.12&18.58&\textbf{30.0}&1.9&5.9&\textbf{25.1}&\textbf{12.1}&\textbf{8.2}&54.9&13.8&32.8&\textbf{6.9}&35.1&8.6&38.3&11.5&14.0&9.6&\textbf{14.4}&\textbf{11.3}\\
CGFormer & 48.07&20.05&29.9&3.4&4.0&17.6&6.8&6.6&\textbf{63.9}&17.2&40.7&5.5&42.7&8.2&38.8&\textbf{24.9}&16.2&17.5&10.2&6.8\\
VLScene & 46.08&19.10&29.0&4.7&7.7&18.3&7.6&7.4&60.1&17.4&39.0&6.0&42.1&9.6&36.5&24.8&17.0&18.8&10.5&6.5\\
Ocean&{48.19}&{20.28}&29.3&3.7&4.6&15.1&7.7&6.8&63.7&17.0&{\textbf{40.9}}&5.0&{43.7}&8.9&{39.2}&24.7&{16.7}&{19.2}&10.8&{8.3}\\

\midrule
VoxDet\textsuperscript{$\dagger$} & 48.32&20.87&29.8&\textbf{4.9}&7.0&17.5&8.1&7.5&62.8&18.3&40.6&5.8&43.7&9.9&38.8&23.5&17.3&\textbf{20.8}&11.2&8.3\\
RayLift (VGGT-$\Omega$) & 48.54&\textbf{21.47}&29.7&4.6&\textbf{8.9}&22.0&10.0&7.8&62.8&18.3&40.8&5.5&44.2&\textbf{10.5}&39.1&24.6&17.4&19.7&11.8&8.9\\
RayLift (MoGe-2) & \textbf{48.82}&\textbf{21.47}&29.8&\textbf{4.9}&\textbf{8.9}&20.2&9.6&7.5&62.9&\textbf{18.8}&40.7&5.6&\textbf{44.3}&\textbf{10.5}&\textbf{39.3}&24.5&\textbf{17.6}&19.6&12.1&9.5\\
\bottomrule
\end{tabular}
\caption{Quantitative results on the SSCBench-KITTI-360 test set. Best values in \textbf{bold}. \(\dagger\) denotes results reproduced using the official code implementation.}
\label{tab:kitti360_main}
\end{table*}
}

\semantickittitable

\subsection{Semantic-Aware Voxel Integrator}

DREL constructs $V_1$ only near the candidate surfaces, whereas the initial voxel representation $V_0$ covers the complete 3D scene. Although both representations occupy the same voxel grid, they differ in spatial support, semantic response, and feature magnitude. Direct addition cannot explicitly account for these differences when determining how ray evidence should update each voxel. We therefore introduce SAVI to integrate the two representations selectively. SAVI restricts $V_1$ to the locations indicated by $M$, adjusts its local contribution according to the semantic weight \(A\) inferred from $V_0$, and aligns the feature magnitudes within the valid region. It also clips extreme updates to prevent individual ray samples from dominating the fused representation. The overall integration is formulated as:
\begin{equation}
\widetilde V
=
V_0+
\mathrm{clip}
\left(
\gamma A\odot M\odot V_1,
-\kappa r_0,\,
\kappa r_0
\right),
\label{eq:savi}
\end{equation}
where $\gamma$ controls the magnitude of $V_1$ relative to $V_0$, \(r_0\) is the root-mean-square magnitude of \(V_0\) over the supported voxels, and \(\kappa\) controls the clipping range. Voxels outside \(M\) remain unchanged, while the supported voxels receive scale-controlled ray evidence. The integrated representation \(\widetilde V\) is subsequently passed to the prediction head for voxel-wise occupancy and semantic prediction.

For semantic guidance, Ray evidence confirms geometric validity for a voxel, yet the update weight assigned to this voxel should not be uniform. To address this, SAVI leverages semantic activations from \(V_0\) to modulate the weight of contributions sourced from \(V_1\). Voxels with prominent occupied-class activations preserve a larger share of ray evidence. By contrast, voxels predicted as free space receive weaker updates, which mitigates adverse effects induced by projection errors.

For feature-scale alignment, $V_0$ and $V_1$ stem from distinct processing branches and may have different feature magnitudes. SAVI controls the root-mean-square magnitude of $V_1$ relative to $V_0$ over valid spatial regions and clips extreme residual responses. This design prevents locally large ray features from dominating the voxel representation.

We retain the training objective of the SSC framework~\cite{li2026voxdet}. The SSC loss
optimizes CoCE and the learnable components of DREL, while SAVI introduces no additional parameters or supervision.

\kittiresulttable

\newcommand{\ablationresulttables}{%
\begin{table*}[!t]
\centering
\small
\begin{minipage}[t]{0.48\textwidth}
\centering
\begin{tabular}{lccccc}
\toprule
Setting & CoCE & DREL & SAVI & IoU & mIoU\\
\midrule
(a) &  &  &  & 47.5281 & 18.7840\\
(b) & $\checkmark$ &  &  & 47.8740 & 18.9462\\
(c) & $\checkmark$ & $\checkmark$ &  & \textbf{48.0343} & 18.9579\\
(d) & $\checkmark$ & $\checkmark$ & $\checkmark$ & 47.9753 & \textbf{19.5366}\\
\bottomrule
\end{tabular}
\captionof{table}{Ablation study of full modules.}
\label{tab:component_ablation}
\end{minipage}
\hfill
\begin{minipage}[t]{0.48\textwidth}
\centering
\begin{tabular}{lcc}
\toprule
Context representation & IoU & mIoU\\
\midrule
DINOv3-2D & 47.7609 & 19.2172\\
Multi-level geometry features & \textbf{47.9753} & \textbf{19.5366}\\
\bottomrule
\end{tabular}
\captionof{table}{Ablation study of context representation schemes.}
\label{tab:context_ablation}
\end{minipage}

\begin{minipage}[t]{0.48\textwidth}
\centering
\begin{tabular}{lcc}
\toprule
DREL configuration & IoU & mIoU\\
\midrule
Full DREL ($8\times8$) & 47.9753 & \textbf{19.5366}\\
Stereo surface cues only & 47.8408 & 18.9270\\
Surface samples only & \textbf{48.0758} & 18.7936\\
$4\times4$ ray cell & 47.9868 & 19.3671\\
w/o voxel reliability weighting & 47.9516 & 19.0594\\
\bottomrule
\end{tabular}
\captionof{table}{Ablation study of DREL module.}
\label{tab:drel_ablation}
\end{minipage}
\hfill
\begin{minipage}[t]{0.48\textwidth}
\centering
\begin{tabular}{lcc}
\toprule
Geometry source & IoU & mIoU\\
\midrule
VGGT-$\Omega$ & 47.9753 & 19.5366\\
MapAnything & 48.0418 & 18.4048\\
Depth Anything 3 & \textbf{48.1069} & 18.9784\\
MoGe-2 & 47.9372 & \textbf{19.6312}\\
\bottomrule
\end{tabular}
\captionof{table}{Ablation study of 3D visual geometry models.}
\label{tab:geometry_source}
\end{minipage}
\end{table*}
}

\ablationresulttables

\section{Experiments}

We evaluate RayLift on SemanticKITTI~\cite{behley2019semantickitti} and SSCBench-KITTI-360~\cite{li2023sscbench}. Additional implementation details and more experiment results are provided in the supplementary material.
\begin{figure*}[!t]
  \centering
  \includegraphics[width=\textwidth]{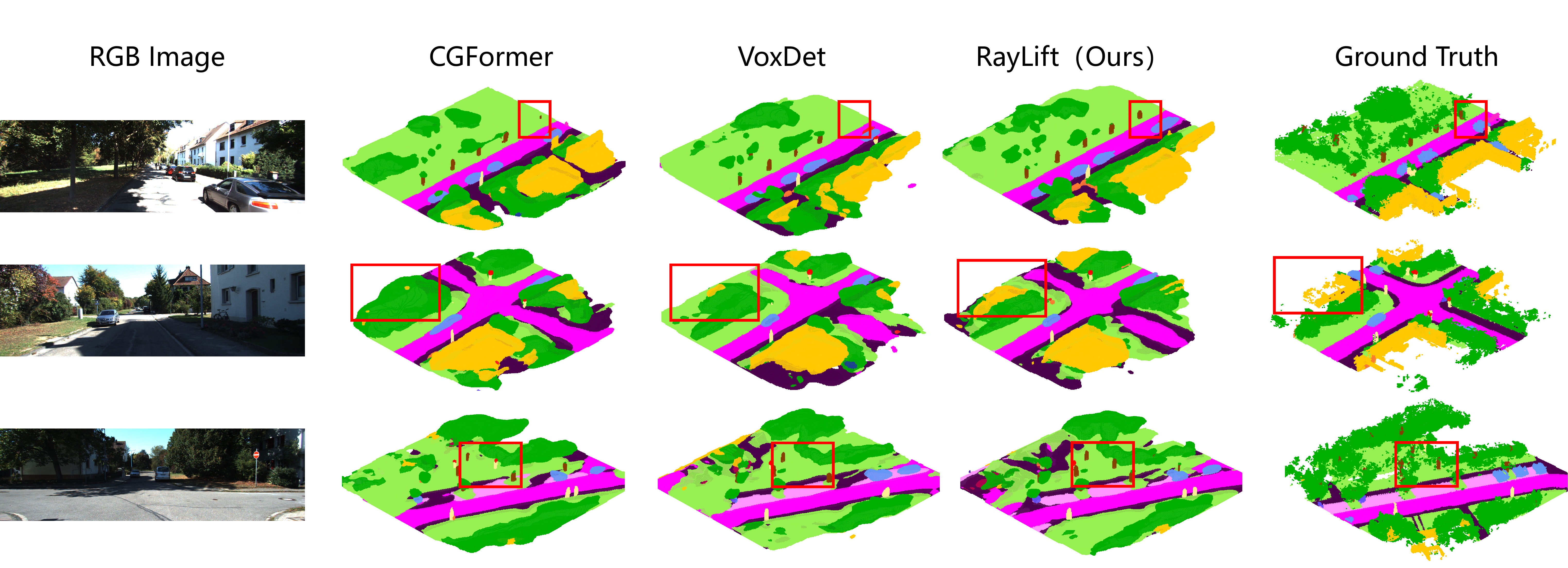}
  \caption{Qualitative comparisons on the SemanticKITTI validation set.}
  \label{fig:qualitative}

  \includegraphics[width=\textwidth]{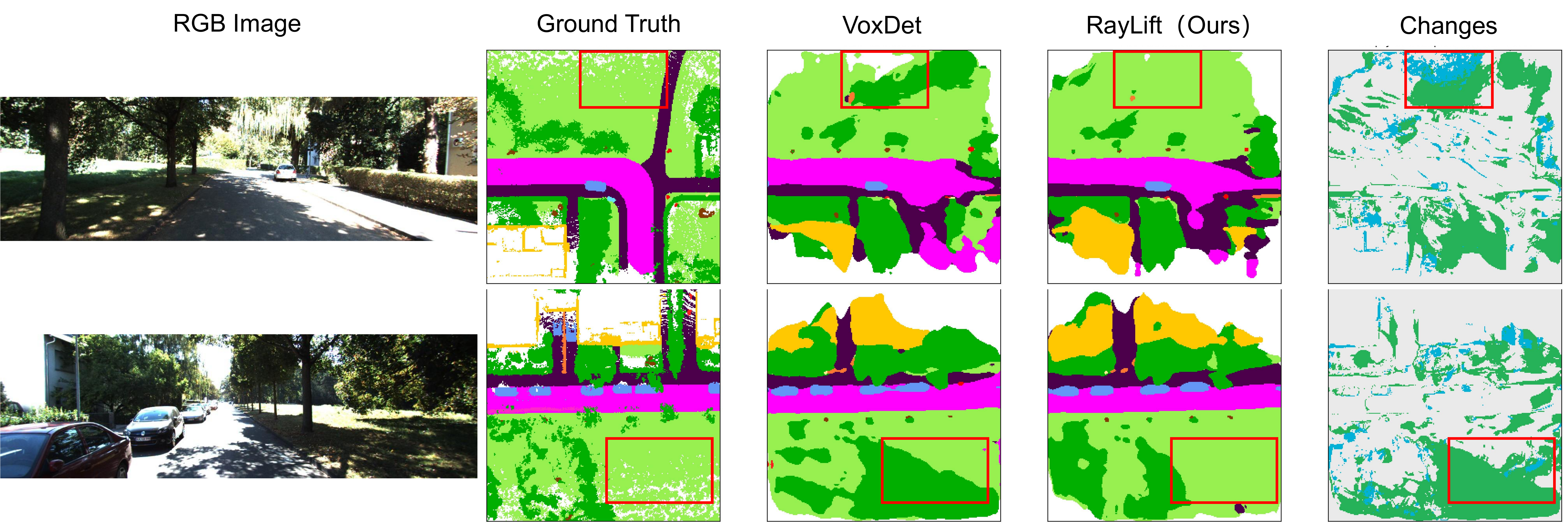}
  \caption{Visualization of prediction corrections on SemanticKITTI. Green indicates semantic corrections in regions previously predicted as occupied, while cyan indicates corrections where occupied voxels were previously predicted as empty.}
  \label{fig:corrections}
\end{figure*}
\subsection{Quantitative Results}

\paragraph{SemanticKITTI Hidden-Test Results.}
Table~\ref{tab:semantickitti_main} reports the results on the SemanticKITTI hidden test set, where RayLift achieves the best overall performance. RayLift with VGGT-$\Omega$~\cite{wang2026vggt} obtains 48.38\% IoU and 18.54\% mIoU, surpassing the reproduced VoxDet~\cite{li2026voxdet} result by 1.10 and 0.54 percentage points, respectively. RayLift with MoGe-2~\cite{wang2026moge} obtains 48.00\% IoU and 18.71\% mIoU, corresponding to gains of 0.72 and 0.71 percentage points. RayLift also achieves the highest accuracy across a broad range of semantic categories, demonstrating stable and effective improvements.

\paragraph{SSCBench-KITTI-360 Results.}
Table~\ref{tab:kitti360_main} presents the results on SSCBench-KITTI-360, where RayLift again achieves the best overall performance. RayLift equipped with VGGT-Omega and MoGe-2 achieves consistent improvements on both metrics. The consistent gains on both datasets further demonstrate the effectiveness and generalization ability of RayLift.

\subsection{Ablation Studies}


\paragraph{Overall Component Ablation.}
Table~\ref{tab:component_ablation} decomposes the contribution of each core component, with VoxDet serving as setting (a). In setting (b), introducing CoCE improves IoU from 47.5281\% to 47.8740\% and mIoU from 18.7840\% to 18.9462\%. Even without DREL, features from the visual geometry model provide the scene context with richer prior information, demonstrating the effectiveness of CoCE. Adding DREL further improves occupancy IoU, indicating that localized ray evidence benefits geometric completion. SAVI subsequently increases mIoU while preserving the occupancy performance, showing that controlled voxel integration converts this evidence into stronger semantic predictions.

\paragraph{CoCE Ablation.}
To select a suitable feature-extraction location in the visual foundation model, Table~\ref{tab:context_ablation} compares multi-level features extracted from the DINOv3~\cite{simeoni2025dinov3} backbone and the geometry aggregator under the same model architecture, fusion scheme, and training setting. This controlled setup ensures a fair comparison between the two feature sources. Although the DINOv3 backbone provides more general 2D object information, the intermediate features from the geometry aggregator yield higher IoU and mIoU and more readily form effective 3D semantic features in our framework.

\paragraph{DREL Ablation.}
Table~\ref{tab:drel_ablation} provides a detailed analysis of ray-evidence construction. Retaining only the stereo surface removes the relative position, geometric discrepancy, and confidence information provided by the second candidate surface. Both IoU and mIoU decrease, and mIoU falls below the result obtained with CoCE alone, showing that the discrepancy between the two candidate surfaces is important to RayLift. In the surface-only setting, we remove the neighboring sample added to each candidate surface. The results show that these two neighboring samples provide useful local context and depth-error information that cannot be retained by the exact surface samples alone. A denser $4\times4$ ray grid does not yield a further gain over the default $8\times8$ setting, indicating that greater ray density does not necessarily improve accuracy; the $8\times8$ cells provide a more suitable balance among spatial coverage, context alignment, and evidence purity. Finally, DREL maps the density of ray evidence received by each voxel to a voxel-level reliability value, which records how much evidence remains after all samples are aggregated and further adjusts the voxel write strength. Removing this reliability weighting decreases mIoU to 19.0594\%, confirming the necessity of this adjustment.

\vspace{-1mm}
\paragraph{3D Vision Model Ablation.}
To compare the effects of different 3D visual foundation models, Table~\ref{tab:geometry_source} evaluates four models by extracting geometric features and monocular depth predictions from comparable network locations. All variants use the same RayLift architecture, parameter settings, and dataset. VGGT-$\Omega$, Depth Anything 3, and MoGe-2 all improve mIoU over the original VoxDet setting, with MoGe-2 achieving the highest mIoU. All four models improve occupancy IoU; Depth Anything 3 achieves the highest IoU but a comparatively lower mIoU, whereas VGGT-$\Omega$ provides a better balance between the two metrics. These results demonstrate that RayLift can effectively use feature and depth priors from different visual geometry models.

\subsection{Qualitative Visualization}

Figure~\ref{fig:qualitative} compares RayLift with CGFormer~\cite{yu2024context} and VoxDet on the SemanticKITTI validation set. For distant objects, RayLift correctly recognizes their semantic categories and recovers their structures; for occluded objects, it reconstructs portions hidden from view. Figure~\ref{fig:corrections} further visualizes the improved regions from a bird's-eye view. RayLift corrects scene structures omitted or misclassified by VoxDet and refines vegetation, terrain, and road boundaries.

\section{Conclusion}

We presented RayLift, a camera-based SSC framework that uses stereo geometry as a metric reference and converts complementary visual geometry priors into localized ray-wise voxel evidence. RayLift constructs geometry-enhanced scene context, organizes candidate surfaces and their reliability along camera rays, and integrates the resulting evidence into the voxel representation before prediction. Experiments on SemanticKITTI and SSCBench-KITTI-360 demonstrate consistent improvements in occupancy IoU and semantic mIoU under matched evaluation settings.

\clearpage
\bibliography{aaai2027}

@InProceedings{Song_2017_CVPR,
author = {Song, Shuran and Yu, Fisher and Zeng, Andy and Chang, Angel X. and Savva, Manolis and Funkhouser, Thomas},
title = {Semantic Scene Completion From a Single Depth Image},
booktitle = {Proceedings of the IEEE Conference on Computer Vision and Pattern Recognition (CVPR)},
month = {July},
year = {2017}
}

@inproceedings{behley2019semantickitti,
  title={Semantickitti: A dataset for semantic scene understanding of lidar sequences},
  author={Behley, Jens and Garbade, Martin and Milioto, Andres and Quenzel, Jan and Behnke, Sven and Stachniss, Cyrill and Gall, Juergen},
  booktitle={2019 IEEE/CVF International Conference on Computer Vision (ICCV)},
  pages={9296--9306},
  year={2019},
  organization={Ieee}
}

@article{li2023sscbench,
  title={SSCBench: Monocular 3D semantic scene completion benchmark in street views},
  author={Li, Yiming and Li, Sihang and Liu, Xinhao and Gong, Moonjun and Li, Kenan and Chen, Nuo and Wang, Zijun and Li, Zhiheng and Jiang, Tao and Yu, Fisher and others},
  year={2023}
}

@inproceedings{cao2022monoscene,
  title={Monoscene: Monocular 3d semantic scene completion},
  author={Cao, Anh-Quan and De Charette, Raoul},
  booktitle={Proceedings of the IEEE/CVF Conference on Computer Vision and Pattern Recognition},
  pages={3991--4001},
  year={2022}
}

@inproceedings{philion2020lift,
  title={Lift, splat, shoot: Encoding images from arbitrary camera rigs by implicitly unprojecting to 3d},
  author={Philion, Jonah and Fidler, Sanja},
  booktitle={European conference on computer vision},
  pages={194--210},
  year={2020},
  organization={Springer}
}

@article{scharstein2002taxonomy,
  title={A taxonomy and evaluation of dense two-frame stereo correspondence algorithms},
  author={Scharstein, Daniel and Szeliski, Richard},
  journal={International journal of computer vision},
  volume={47},
  number={1},
  pages={7--42},
  year={2002},
  publisher={Springer}
}

@inproceedings{wang2025vggt,
  title={Vggt: Visual geometry grounded transformer},
  author={Wang, Jianyuan and Chen, Minghao and Karaev, Nikita and Vedaldi, Andrea and Rupprecht, Christian and Novotny, David},
  booktitle={Proceedings of the Computer Vision and Pattern Recognition Conference},
  pages={5294--5306},
  year={2025}
}

@InProceedings{Schonberger_2016_CVPR,
author = {Schonberger, Johannes L. and Frahm, Jan-Michael},
title = {Structure-From-Motion Revisited},
booktitle = {Proceedings of the IEEE Conference on Computer Vision and Pattern Recognition (CVPR)},
month = {June},
year = {2016}
}

@inproceedings{wang2024dust3r,
  title={Dust3r: Geometric 3d vision made easy},
  author={Wang, Shuzhe and Leroy, Vincent and Cabon, Yohann and Chidlovskii, Boris and Revaud, Jerome},
  booktitle={Proceedings of the IEEE/CVF conference on computer vision and pattern recognition},
  pages={20697--20709},
  year={2024}
}

@inproceedings{leroy2024grounding,
  title={Grounding image matching in 3d with mast3r},
  author={Leroy, Vincent and Cabon, Yohann and Revaud, J{\'e}r{\^o}me},
  booktitle={European conference on computer vision},
  pages={71--91},
  year={2024},
  organization={Springer}
}

@inproceedings{yang2024depth,
  title={Depth anything: Unleashing the power of large-scale unlabeled data},
  author={Yang, Lihe and Kang, Bingyi and Huang, Zilong and Xu, Xiaogang and Feng, Jiashi and Zhao, Hengshuang},
  booktitle={Proceedings of the IEEE/CVF conference on computer vision and pattern recognition},
  pages={10371--10381},
  year={2024}
}

@article{lin2025depth,
  title={Depth anything 3: Recovering the visual space from any views},
  author={Lin, Haotong and Chen, Sili and Liew, Junhao and Chen, Donny Y and Li, Zhenyu and Shi, Guang and Feng, Jiashi and Kang, Bingyi},
  journal={arXiv preprint arXiv:2511.10647},
  year={2025}
}

@inproceedings{keetha2026mapanything,
  title={Mapanything: Universal feed-forward metric 3d reconstruction},
  author={Keetha, Nikhil Varma and M{\"u}ller, Norman and Sch{\"o}nberger, Johannes and Porzi, Lorenzo and Zhang, Yuchen and Fischer, Tobias and Knapitsch, Arno and Zauss, Duncan and Weber, Ethan and Antunes, Nelson and others},
  booktitle={Thirteenth International Conference on 3D Vision},
  year={2026}
}

@article{wang2026moge,
  title={Moge-2: Accurate monocular geometry with metric scale and sharp details},
  author={Wang, Ruicheng and Xu, Sicheng and Dong, Yue and Deng, Yu and Xiang, Jianfeng and Lv, Zelong and Sun, Guangzhong and Tong, Xin and Yang, Jiaolong},
  journal={Advances in Neural Information Processing Systems},
  volume={38},
  pages={35928--35959},
  year={2026}
}

@article{wang2026vggt,
  title   = {{VGGT}-$\Omega$},
  author  = {Wang, Jianyuan and Chen, Minghao and Zhang, Shangzhan and Karaev, Nikita and Sch{\"o}nberger, Johannes and Labatut, Patrick and Bojanowski, Piotr and Novotny, David and Vedaldi, Andrea and Rupprecht, Christian},
  journal = {arXiv preprint arXiv:2605.15195},
  year    = {2026}
}

@inproceedings{huang2023tri,
  title={Tri-perspective view for vision-based 3d semantic occupancy prediction},
  author={Huang, Yuanhui and Zheng, Wenzhao and Zhang, Yunpeng and Zhou, Jie and Lu, Jiwen},
  booktitle={Proceedings of the IEEE/CVF conference on computer vision and pattern recognition},
  pages={9223--9232},
  year={2023}
}

@inproceedings{li2023voxformer,
  title={Voxformer: Sparse voxel transformer for camera-based 3d semantic scene completion},
  author={Li, Yiming and Yu, Zhiding and Choy, Christopher and Xiao, Chaowei and Alvarez, Jose M and Fidler, Sanja and Feng, Chen and Anandkumar, Anima},
  booktitle={Proceedings of the IEEE/CVF conference on computer vision and pattern recognition},
  pages={9087--9098},
  year={2023}
}

@inproceedings{zhang2023occformer,
  title={Occformer: Dual-path transformer for vision-based 3d semantic occupancy prediction},
  author={Zhang, Yunpeng and Zhu, Zheng and Du, Dalong},
  booktitle={Proceedings of the IEEE/CVF International Conference on Computer Vision},
  pages={9433--9443},
  year={2023}
}

@article{li2023bridging,
  title={Bridging stereo geometry and BEV representation with reliable mutual interaction for semantic scene completion},
  author={Li, Bohan and Sun, Yasheng and Liang, Zhujin and Du, Dalong and Zhang, Zhuanghui and Wang, Xiaofeng and Wang, Yunnan and Jin, Xin and Zeng, Wenjun},
  journal={arXiv preprint arXiv:2303.13959},
  year={2023}
}

@inproceedings{yao2025depthssc,
  title={DepthSSC: Monocular 3D semantic scene completion via depth-spatial alignment and voxel adaptation},
  author={Yao, Jiawei and Zhang, Jusheng and Pan, Xiaochao and Wu, Tong and Xiao, Canran},
  booktitle={2025 IEEE/CVF Winter Conference on Applications of Computer Vision (WACV)},
  pages={2154--2163},
  year={2025},
  organization={IEEE}
}

@article{yu2024context,
  title={Context and geometry aware voxel transformer for semantic scene completion},
  author={Yu, Zhu and Zhang, Runmin and Ying, Jiacheng and Yu, Junchen and Hu, Xiaohai and Luo, Lun and Cao, Si-Yuan and Shen, Hui-Liang},
  journal={Advances in Neural Information Processing Systems},
  volume={37},
  pages={1531--1555},
  year={2024}
}

@inproceedings{wang2024not,
  title={Not all voxels are equal: Hardness-aware semantic scene completion with self-distillation},
  author={Wang, Song and Yu, Jiawei and Li, Wentong and Liu, Wenyu and Liu, Xiaolu and Chen, Junbo and Zhu, Jianke},
  booktitle={Proceedings of the IEEE/CVF Conference on Computer Vision and Pattern Recognition},
  pages={14792--14801},
  year={2024}
}

@article{xiao2024instance,
  title={Instance-aware monocular 3d semantic scene completion},
  author={Xiao, Haihong and Xu, Hongbin and Kang, Wenxiong and Li, Yuqiong},
  journal={IEEE Transactions on Intelligent Transportation Systems},
  volume={25},
  number={7},
  pages={6543--6554},
  year={2024},
  publisher={IEEE}
}

@article{jiang2023symphonize,
  title={Symphonize 3d semantic scene completion with contextual instance queries},
  author={Jiang, Haoyi and Cheng, Tianheng and Gao, Naiyu and Zhang, Haoyang and Lin, Tianwei and Liu, Wenyu and Wang, Xinggang},
  journal={arXiv preprint arXiv:2306.15670},
  year={2023}
}

@inproceedings{wang2025vlscene,
  title={VLScene: Vision-language guidance distillation for camera-based 3D semantic scene completion},
  author={Wang, Meng and Pi, Huilong and Li, Ruihui and Qin, Yunchuan and Tang, Zhuo and Li, Kenli},
  booktitle={Proceedings of the AAAI Conference on Artificial Intelligence},
  volume={39},
  number={8},
  pages={7808--7816},
  year={2025}
}

@inproceedings{guo2025sgformer,
  title={Sgformer: Satellite-ground fusion for 3d semantic scene completion},
  author={Guo, Xiyue and Hu, Jiarui and Hu, Junjie and Bao, Hujun and Zhang, Guofeng},
  booktitle={Proceedings of the IEEE/CVF Conference on Computer Vision and Pattern Recognition},
  pages={11929--11938},
  year={2025}
}

@article{li2026voxdet,
  title={Voxdet: Rethinking 3d semantic scene completion as dense object detection},
  author={Li, Wuyang and Yu, Zhu and Alahi, Alexandre},
  journal={Advances in Neural Information Processing Systems},
  volume={38},
  pages={81004--81038},
  year={2026}
}

@article{simeoni2025dinov3,
  title={Dinov3},
  author={Sim{\'e}oni, Oriane and Vo, Huy V and Seitzer, Maximilian and Baldassarre, Federico and Oquab, Maxime and Jose, Cijo and Khalidov, Vasil and Szafraniec, Marc and Yi, Seungeun and Ramamonjisoa, Micha{\"e}l and others},
  journal={arXiv preprint arXiv:2508.10104},
  year={2025}
}

@inproceedings{wang2026objectcentric,
  title={Towards 3D object-centric feature learning for semantic scene completion},
  author={Wang, Weihua and Cui, Yubo and Lin, Xiangru and Li, Zhiheng and Fang, Zheng},
  booktitle={Proceedings of the AAAI Conference on Artificial Intelligence},
  volume={40},
  number={12},
  pages={10136--10144},
  year={2026}
}

@article{wen2025fast,
  title={Fast-FoundationStereo: Real-Time Zero-Shot Stereo Matching},
  author={Wen, Bowen and Dewan, Shaurya and Birchfield, Stan},
  journal={arXiv preprint arXiv:2512.11130},
  year={2025}
}

@inproceedings{shamsafar2022mobilestereonet,
  title={Mobilestereonet: Towards lightweight deep networks for stereo matching},
  author={Shamsafar, Faranak and Woerz, Samuel and Rahim, Rafia and Zell, Andreas},
  booktitle={Proceedings of the ieee/cvf winter conference on applications of computer vision},
  pages={2417--2426},
  year={2022}
}
\clearpage
\appendix
\setcounter{topnumber}{3}
\setcounter{totalnumber}{5}
\setcounter{figure}{0}
\setcounter{table}{0}
\setcounter{equation}{0}

\section{Appendix/Supplemental Material}

This appendix provides additional implementation details and more comprehensive experimental results.

\subsection{Experimental Setup}

\paragraph{Datasets.} We evaluate RayLift on two widely used public benchmarks, SemanticKITTI~\cite{behley2019semantickitti} and SSCBench-KITTI-360~\cite{li2023sscbench}. SemanticKITTI contains images at a resolution of \(1226\times370\) and comprises 22 sequences (00--21). Sequences 00--07, 09, and 10 are used for training, sequence 08 for validation, and sequences 11--21 for online evaluation on the hidden test server. The benchmark defines 20 labels, including 19 semantic classes and one empty class. SSCBench-KITTI-360 contains images at a resolution of \(1408\times376\) and comprises nine sequences, of which seven are used for training, one for validation, and one for final testing. Each voxel is assigned one of 19 labels, including 18 semantic classes and one empty class.

For both datasets, the observed frustum extends \(51.2\,\mathrm{m}\) forward, \(25.6\,\mathrm{m}\) to either side, and \(6.4\,\mathrm{m}\) vertically. The final scene prediction has a voxel resolution of \(256\times256\times32\), with a voxel size of \(0.2\,\mathrm{m}\).

\paragraph{Evaluation Metrics.} Following established practice, we evaluate geometric completion and semantic completion using intersection over union (IoU) and mean intersection over union (mIoU), respectively. IoU measures occupied-voxel reconstruction, whereas mIoU measures voxel-wise semantic prediction.

\subsection{Implementation Details}

\subsubsection{Complementary Context Encoder}

The input images are resized to \(384\times1280\) on SemanticKITTI and \(384\times1408\) on SSCBench-KITTI-360, and the camera projection transformations are updated accordingly to match the CoCE input size. CoCE extracts image features and visual-geometry depth from the left image and obtains stereo depth from the stereo pair. We use VGGT-\(\Omega\) as the 3D vision foundation model and extract its intermediate geometry-aggregator features and predicted depth offline. Four feature layers are used. For each layer, VGGT-\(\Omega\) produces features at a spatial stride of \(1/14\); these features are normalized, mapped through a fixed projection, and interpolated to one-eighth of the input-image resolution. The four layers are then concatenated along the channel dimension to obtain the visual-geometry feature \(\boldsymbol{F}_{\mathrm{vfm}}\), which has the same spatial size and number of channels as the 2D image feature.

\subsubsection{Depth Ray Evidence Lifter}

Because depth predicted by the 3D vision foundation model is not necessarily metric, it cannot be directly combined with stereo depth to construct ray evidence. We therefore calibrate its scale on a per-frame basis using stereo depth. For pixels at which both estimates are finite and positive, we compute the ratio between stereo depth and visual-geometry depth. After discarding the lowest and highest 5\% of the ratio distribution, the median of the remaining ratios is used as the scale factor. Calibration is performed only when a frame contains at least 100 valid overlapping pixels.

Let \(r\) denote a ray cell. The definitions and interpretations of the descriptor components are given in Table~\ref{tab:ray_descriptor}.

\begin{table*}[t]
\centering
\small
\setlength{\tabcolsep}{3pt}
\renewcommand{\arraystretch}{1.45}
\begin{tabular}{@{}p{0.32\textwidth}p{0.29\textwidth}p{0.33\textwidth}@{}}
\toprule
\centering\textbf{Formula} & \centering\textbf{Related Definition} & \centering\arraybackslash\textbf{Interpretation}\\
\midrule
\vcell{\centering\(\displaystyle h_r=-\frac{1}{\log N_d}\sum_{j=1}^{N_d}P_r(j)\log(P_r(j))\)\par} & \vcell{\(P_r(j)\) denotes the probability of ray cell \(r\) in the \(j\)-th depth interval, and \(N_d\) is the number of depth intervals.} & \vcell{Depth-distribution entropy. A smaller \(h_r\) indicates a concentrated probability distribution and a more definite depth location from the original lifting path; a larger \(h_r\) indicates that the probability is dispersed along the ray and that the surface location is more uncertain.}\\[-\rowheight]
\printcellmiddle & \printcelltop & \printcelltop\\[2pt]
\vcell{\centering\(\displaystyle \delta_r=\operatorname{clip}\!\left(\log\frac{d_r^g}{d_r^s},-2,2\right)\)\par} & \vcell{\(d_r^s\) and \(d_r^g\) denote the stereo depth and scale-corrected visual-geometry depth, respectively.} & \vcell{Signed depth discrepancy. When \(\delta_r>0\), the visual-geometry surface is behind the stereo surface; when \(\delta_r<0\), it is in front of the stereo surface; \(\delta_r=0\) indicates that the two surfaces are close or that the candidate pair is invalid.}\\[-\rowheight]
\printcellmiddle & \printcelltop & \printcelltop\\[2pt]
\vcell{\centering\(\displaystyle \lVert\delta_r\rVert=\operatorname{abs}(\delta_r)\)\par} & \vcell{The absolute value of \(\delta_r\).} & \vcell{Unsigned depth discrepancy. A small value indicates that the two surfaces are close, whereas a large value indicates a clear geometric discrepancy between them.}\\[-\rowheight]
\printcellmiddle & \printcelltop & \printcelltop\\[2pt]
\vcell{\centering\(\displaystyle \begin{gathered}\bar c_r^g=\log\!\left(1+\max(c_r^g,0)\right),\\[-1pt]c_r=\operatorname{clip}\!\left(\frac{\bar c_r^g-\mu_c}{\sqrt{\max(\sigma_c^2,10^{-6})}},-5,5\right)\end{gathered}\)\par} & \vcell{\(c_r^g\) denotes the raw visual-geometry confidence, while \(\mu_c\) and \(\sigma_c^2\) are its mean and variance over the valid ray cells in the current image.} & \vcell{Relative confidence. \(c_r>0\) indicates that the candidate surface has higher confidence than other valid ray-cell locations in the current image.}\\[-\rowheight]
\printcellmiddle & \printcelltop & \printcelltop\\[2pt]
\vcell{\centering\(\displaystyle v_r=\begin{cases}1, & d_r^s,d_r^g\in[2,58),\ \bar c_r^g>0,\\0, & \text{otherwise}.\end{cases}\)\par} & \vcell{\(d_r^s\) and \(d_r^g\) denote the stereo depth and scale-corrected visual-geometry depth, respectively; the valid depth range is \([2,58)\,\mathrm{m}\).} & \vcell{\(v_r\) is a binary indicator of whether the stereo depth, visual-geometry depth, and confidence jointly form a valid candidate-surface pair.}\\[-\rowheight]
\printcellmiddle & \printcelltop & \printcelltop\\[2pt]
\vcell{\centering\(\displaystyle e_r=\operatorname{clip}\!\left(\frac{d_r^{\max}-d_r^{\min}}{\max(d_r^{\min},10^{-3})},0,2\right)\)\par} & \vcell{\(d_r^{\max}\) and \(d_r^{\min}\) denote the maximum and minimum depth values within the ray cell, respectively.} & \vcell{Within-cell depth variation. A smaller \(e_r\) indicates a relatively smooth surface within the cell, whereas a larger \(e_r\) indicates greater depth variation and may correspond to an object boundary.}\\[-\rowheight]
\printcellmiddle & \printcelltop & \printcelltop\\
\bottomrule
\end{tabular}
\caption{Formulas, related definitions, and interpretations of the ray geometry descriptor components.}
\label{tab:ray_descriptor}
\end{table*}

After ray sampling, each sample is represented by a 3D position \(x_{r,k}\), a context feature \(f_{r,k}\), and a write weight \(\widehat{\omega}_{r,k}\). We integrate these samples into the voxel grid. Valid samples are mapped to the \(128\times128\times16\) internal feature grid of the prediction head, which covers \(x\in[0,51.2)\), \(y\in[-25.6,25.6)\), and \(z\in[-2,4.4)\,\mathrm{m}\), with a voxel size of \(0.4\,\mathrm{m}\). According to its continuous coordinate and distance to neighboring voxel centers, each sample is assigned to at most eight neighboring voxels, with contributions determined jointly by trilinear interpolation coefficients and the sample write weight. When multiple samples fall into the same voxel, their features are aggregated with these weights while the accumulated evidence weight \(\rho\) is recorded. The aggregated feature is normalized by \(\rho\) so that its magnitude does not grow directly with the number of samples. For samples near the grid boundary, contributions outside the scene range are discarded and the remaining valid contributions are retained. Samples that lie entirely outside the scene or have invalid depth produce no voxel evidence, and voxels that receive no valid sample remain zero. After aggregation, each voxel contains a voxel feature and \(\rho\). We map \(\rho\) to a voxel-level weight and use a threshold of \(0.01\) to obtain the valid-position mask \(M\), which distinguishes voxels that actually receive ray evidence from uncovered regions. The voxel feature and \(\rho\) are then concatenated along the channel dimension and processed by a lightweight 3D convolutional encoder, which organizes discrete features from different rays and candidate surfaces within valid voxels. The encoded result is restricted by \(M\), modulated by the voxel weight, and globally scaled to produce the local ray-wise voxel evidence \(V_1\).

\subsubsection{Semantic-Aware Voxel Integrator}

We compute the semantic weight \(A\) and global scaling coefficient \(\gamma\) as follows. The existing semantic transformation of the prediction head maps \(V_0\) to the pre-integration semantic response \(Z^0\). For voxel \(v\), the non-empty response difference is defined as
\begin{equation}
m_v=\max_{c>0}Z^0_{v,c}-Z^0_{v,0}.
\label{eq:supp_nonempty_response}
\end{equation}
When \(m_v\) is large, the voxel already contains a clear non-empty scene response, allowing the ray evidence to further refine its surface structure or semantic feature. When \(m_v\) is small, the location is more likely to represent free space, and the write strength is therefore reduced. The semantic weight is defined as
\begin{equation}
A=\max\!\left(a_{\min},\sigma\!\left(\frac{m_v+\tau}{T}\right)\right),
\label{eq:supp_semantic_weight}
\end{equation}
where \(\tau\) controls the response center, \(T\) controls the smoothness of the weight transition, and \(a_{\min}\) retains a minimum update strength for voxels with valid ray support. We set these values to \(0.75\), \(0.35\), and \(0.02\), respectively.

For the global scaling coefficient \(\gamma\), the root-mean-square magnitudes of \(V_1\) and \(V_0\) over the region indicated by \(M\) are defined as
\begin{equation}
r_1=\sqrt{\frac{\sum_vM_v\lVert V_{1,v}\rVert_2^2}{C_{\mathrm{vox}}\sum_vM_v}},\;r_0=\sqrt{\frac{\sum_vM_v\lVert V_{0,v}\rVert_2^2}{C_{\mathrm{vox}}\sum_vM_v}}.
\label{eq:supp_rms}
\end{equation}
The global scaling coefficient for the ray update is then
\begin{equation}
\gamma=\operatorname{clip}\!\left(\frac{0.1r_0}{r_1},0,1\right).
\label{eq:supp_scale}
\end{equation}
This design attenuates the auxiliary ray features only when they are excessively strong, without actively amplifying weak ray responses. 
Ultimately, the effective contribution of the ray evidence is jointly controlled by the ray-support mask \(M_v\), the voxel-wise semantic weight \(A_v\), and the global scaling coefficient \(\gamma\). Specifically, \(M_v\) determines which voxels can receive auxiliary evidence, \(A_v\) adjusts the relative update strength of each supported voxel according to its local semantic state, and \(\gamma\) constrains the overall magnitude of the ray branch relative to the original voxel representation.
\subsubsection{Training Configuration}

RayLift is trained on two NVIDIA A6000 GPUs with a batch size of two per GPU and a total batch size of four. We use AdamW with a weight decay of \(0.01\). On SemanticKITTI, the initial learning rate is \(3\times10^{-4}\) and the model is trained for 25,000 iterations; on SSCBench-KITTI-360, the initial learning rate is \(2\times10^{-4}\) and the model is trained for 27,000 iterations. The OneCycle schedule linearly increases the learning rate to its preset peak during the first 5\% of the training steps and applies cosine decay thereafter. Gradients are clipped with a maximum norm of 20. All formal experiments are initialized from the same pretrained weights.

\subsection{Quantitative Results}

\noindent\textbf{SemanticKITTI Validation Results.}\enspace Table~\ref{tab:semantickitti_val} reports additional quantitative results on the SemanticKITTI validation set. RayLift achieves the best IoU and mIoU among the listed methods.

\begin{table*}[t]
\centering
\small
\setlength{\tabcolsep}{1.1pt}
\renewcommand{\arraystretch}{1.05}
\begin{tabular}{@{}lcc*{19}{c}@{}}
\toprule
Method & IoU & mIoU & \rotatebox{90}{road} & \rotatebox{90}{sidewalk} & \rotatebox{90}{parking} & \rotatebox{90}{other-grnd.} & \rotatebox{90}{building} & \rotatebox{90}{car} & \rotatebox{90}{truck} & \rotatebox{90}{bicycle} & \rotatebox{90}{motorcycle} & \rotatebox{90}{other-veh.} & \rotatebox{90}{vegetation} & \rotatebox{90}{trunk} & \rotatebox{90}{terrain} & \rotatebox{90}{person} & \rotatebox{90}{bicyclist} & \rotatebox{90}{motorcyclist} & \rotatebox{90}{fence} & \rotatebox{90}{pole} & \rotatebox{90}{traf.-sign}\\
\midrule
MonoScene & 36.86 & 11.08 & 56.52&26.72&14.27&0.46&14.09&23.26&6.98&0.61&0.45&1.48&17.89&2.81&29.64&1.86&1.20&0.00&5.84&4.14&2.25\\
VoxFormer & 44.15 & 13.35 & 53.57&26.52&19.69&0.42&19.54&26.54&7.26&1.28&0.56&7.81&26.10&6.10&33.06&1.93&1.97&0.00&7.31&9.15&4.94\\
OccFormer & 36.50 & 13.46 & 58.85&26.88&19.61&0.31&14.40&25.09&25.53&0.81&1.19&8.52&19.63&3.93&32.62&2.78&2.82&0.00&5.61&4.26&2.86\\
Symphonize & 41.92 & 14.89 & 56.37&27.58&15.28&\textbf{0.95}&21.64&28.68&20.44&2.54&2.82&13.89&25.72&6.60&30.87&3.52&2.24&0.00&8.40&9.57&5.76\\
CGFormer & 45.99 & 16.87 & 65.51&32.31&20.82&0.16&23.52&34.32&19.44&4.61&2.71&7.67&26.93&8.83&39.54&2.38&4.08&0.00&9.20&10.67&7.84\\
VLScene & 44.69 & 17.83 & 63.10&31.10&\textbf{24.40}&0.20&24.90&33.40&30.70&1.80&3.60&18.30&26.00&8.10&35.30&4.30&2.60&0.00&12.10&11.90&6.30\\
Ocean & 46.40 & 17.39 & 66.1&34.3&21.9&0.1&23.4&34.0&19.3&3.2&2.0&15.3&28.1&8.9&40.0&3.7&1.2&0.0&10.3&10.8&7.8\\
\midrule
VoxDet\textsuperscript{$\dagger$} & 47.53 & 18.78 & 66.0&\textbf{35.6}&23.3&0.2&26.6&34.4&25.5&4.1&5.3&14.6&\textbf{29.7}&10.2&41.3&4.3&3.2&0.0&11.1&12.8&8.8\\
RayLift (VGGT-\(\Omega\)) & \textbf{47.98} & 19.54 & 66.3&34.5&22.3&0.2&\textbf{26.8}&\textbf{34.7}&33.7&\textbf{6.3}&\textbf{7.6}&14.2&29.4&\textbf{10.6}&41.9&4.3&\textbf{4.8}&0.0&12.3&12.6&8.7\\
RayLift (MoGe-2) & 47.94 & \textbf{19.63} & \textbf{66.6}&34.4&21.4&0.1&26.5&\textbf{34.7}&\textbf{34.5}&5.6&4.8&\textbf{19.3}&29.2&10.0&\textbf{42.1}&\textbf{4.8}&3.3&0.0&\textbf{13.4}&\textbf{13.2}&\textbf{9.1}\\
\bottomrule
\end{tabular}
\caption{Quantitative results on the SemanticKITTI validation set. The best overall and class-wise results are shown in bold. \(\dagger\) denotes our reproduction under the same experimental setting as RayLift.}
\label{tab:semantickitti_val}
\end{table*}

\noindent\textbf{Stereo Depth Source.}\enspace Table~\ref{tab:stereo_source} compares two stereo models used to provide the metric reference for RayLift, MSNet~\cite{shamsafar2022mobilestereonet} and Fast-FoundationStereo~\cite{wen2025fast}. Compared with MSNet, Fast-FoundationStereo improves validation IoU and mIoU by 0.27 and 0.21 percentage points, respectively. On the hidden test set, however, MSNet achieves 0.60 points higher IoU and 0.44 points higher mIoU. Both depth sources therefore support the construction of effective ray evidence, indicating that RayLift is not tied to a particular stereo estimator. The reversal between the validation and hidden-test rankings further shows that cross-sequence generalization is not determined solely by the accuracy of the stereo reference.\newline
\textbf{Geometry Integration Strategy.}\enspace As shown in Table~\ref{tab:integration_strategy}, explicitly retaining the candidate surfaces provided by different depth sources as local ray-wise voxel evidence achieves better semantic completion performance than calibrating the additional geometry and directly integrating it into the LSS depth distribution. This result supports our motivation for preserving heterogeneous depth predictions as distinct surface hypotheses rather than collapsing them into a single lifting distribution.

\begin{table}[t]
\centering
\small
\setlength{\tabcolsep}{4.2pt}
\begin{tabular}{@{}lcccc@{}}
\toprule
Stereo reference & \multicolumn{2}{c}{Validation} & \multicolumn{2}{c}{Hidden test}\\
\cmidrule(lr){2-3}\cmidrule(l){4-5}
& IoU & mIoU & IoU & mIoU\\
\midrule
MSNet & 47.98 & 19.54 & \textbf{48.38} & \textbf{18.54}\\
Fast-FoundationStereo & \textbf{48.25} & \textbf{19.75} & 47.78 & 18.10\\
\bottomrule
\end{tabular}
\caption{Effect of the stereo-depth source on SemanticKITTI. Validation results are taken at each model's best-mIoU checkpoint, while hidden-test results are obtained from the online evaluation server. All values are percentages.}
\label{tab:stereo_source}
\end{table}

\begin{table}[t]
\centering
\small
\setlength{\tabcolsep}{3pt}
\begin{tabular}{@{}p{0.60\columnwidth}cc@{}}
\toprule
Method & IoU & mIoU\\
\midrule
Modify Depth Probability Distribution & 47.8333 & 19.0373\\
RayLift & \textbf{47.9753} & \textbf{19.5366}\\
\bottomrule
\end{tabular}
\caption{Comparison of geometry-prior integration strategies on the SemanticKITTI validation set. All values are percentages.}
\label{tab:integration_strategy}
\end{table}

\setcounter{figure}{0}
\begin{figure}[!t]
\centering
\includegraphics[width=0.94\columnwidth]{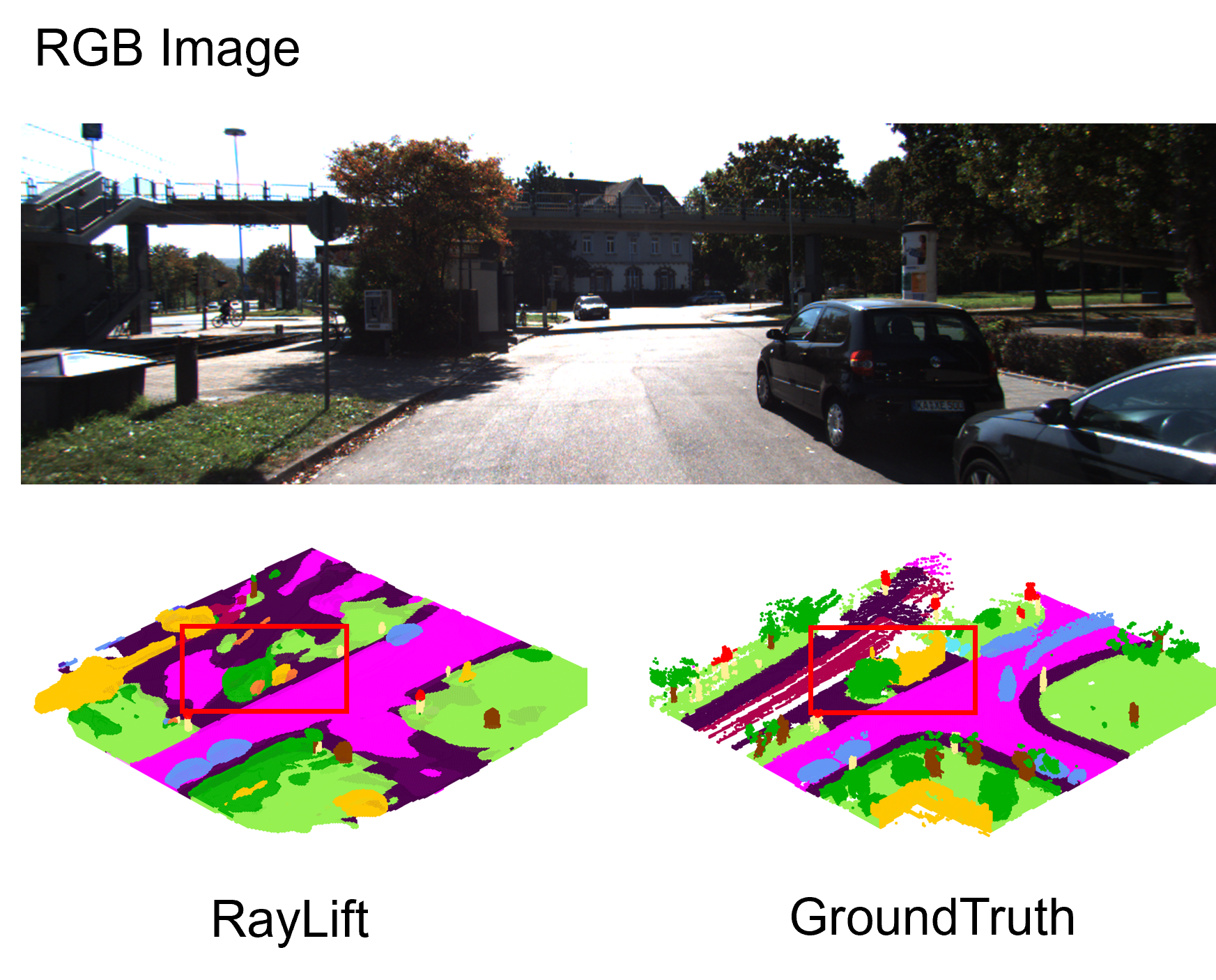}
\caption{Failure Cases.}
\label{fig:supp_failure}
\end{figure}

\subsection{Qualitative Results}
\begin{figure*}[p]
\centering
\includegraphics[width=0.90\textwidth]{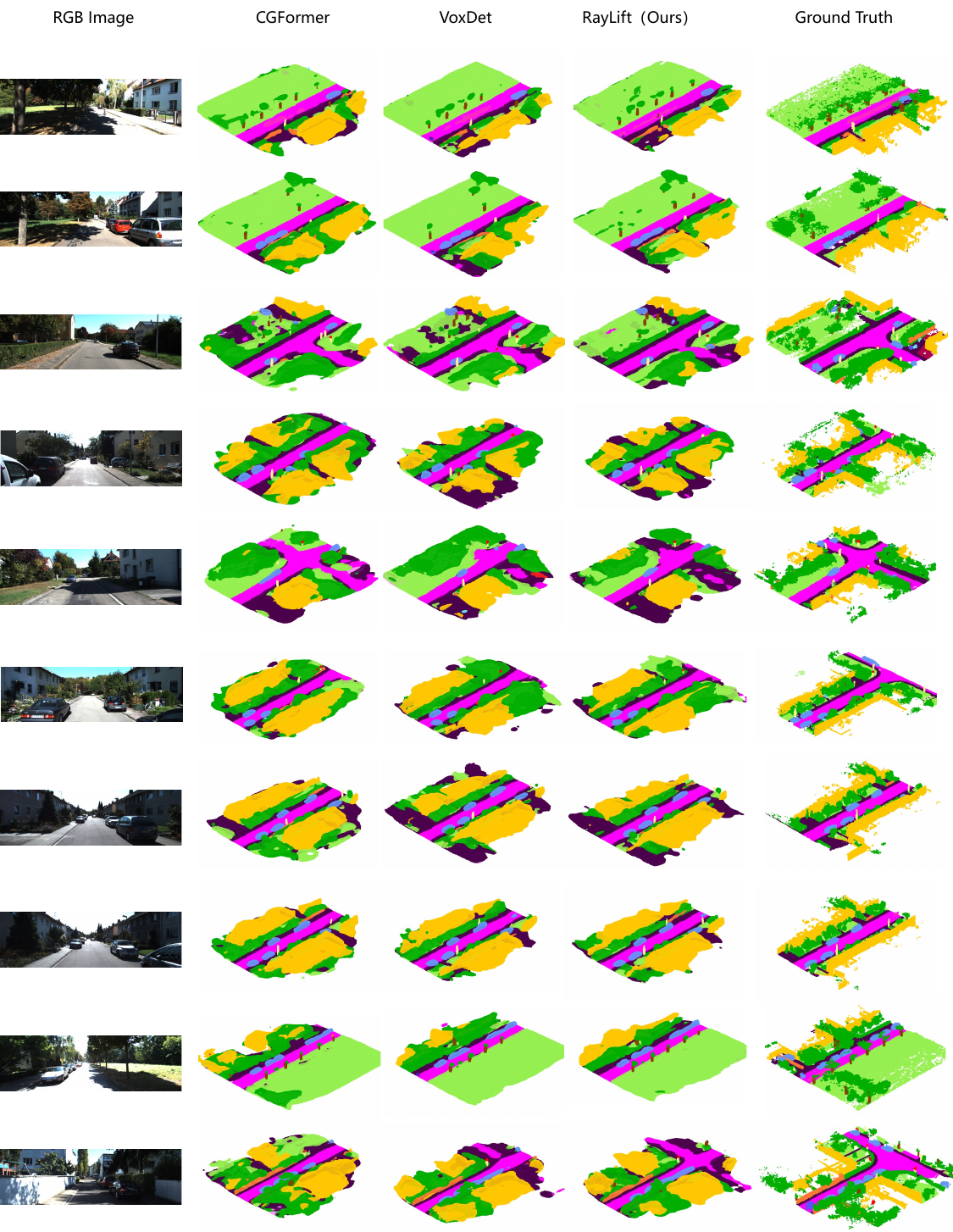}
\caption{Additional qualitative comparisons on the SemanticKITTI validation set. From left to right: input RGB image, CGFormer, VoxDet, RayLift, and ground truth.}
\label{fig:supp_qualitative}
\end{figure*}

\begin{figure*}[p]
\centering
\includegraphics[width=0.848\textwidth]{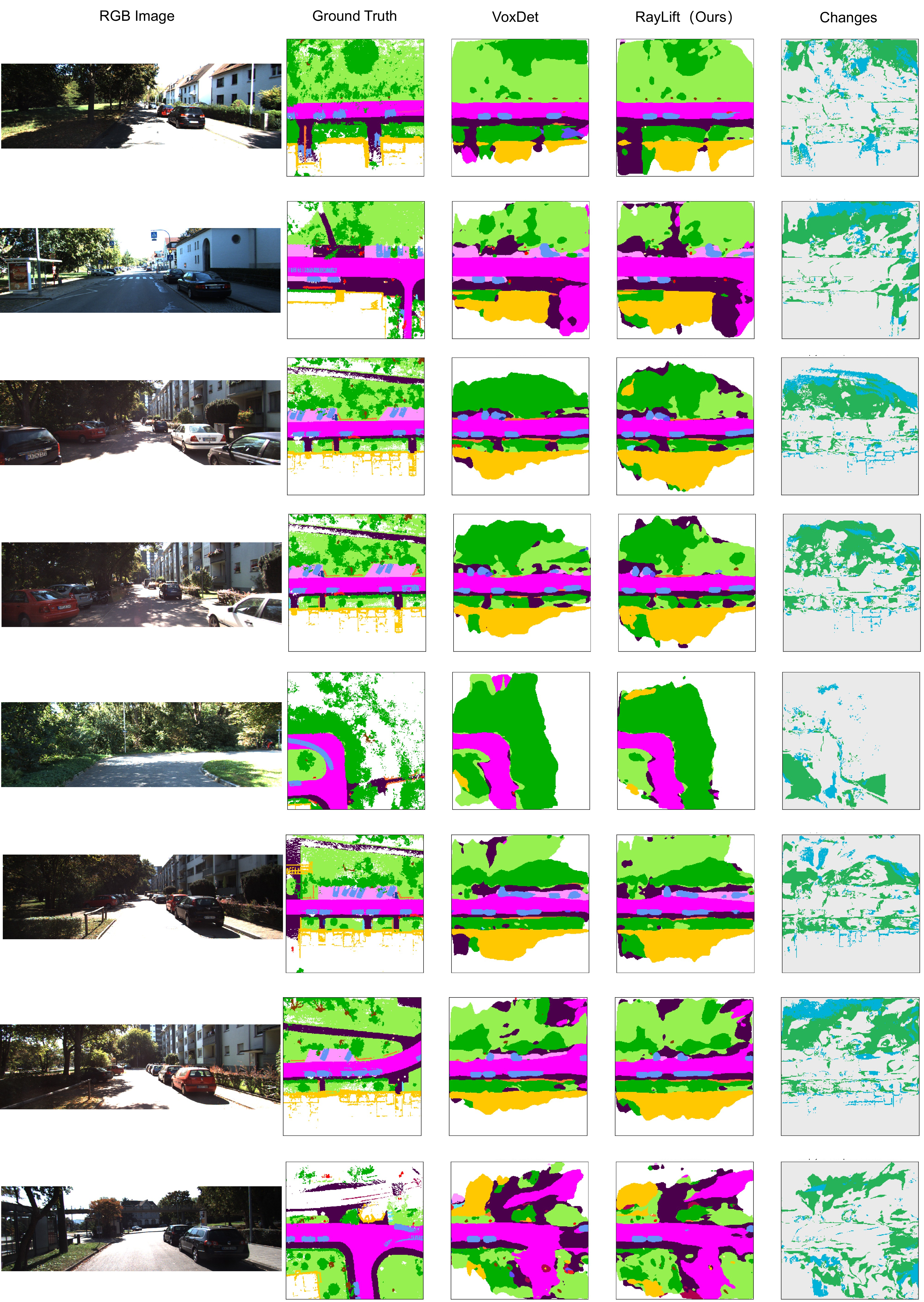}
\caption{Additional qualitative visualization of the corrected regions on the SemanticKITTI validation set. From left to right: input RGB image, ground truth, VoxDet, RayLift, and corrected regions. Green denotes corrected semantic labels in previously occupied voxels, while cyan denotes occupied voxels recovered from previous free-space predictions.}
\label{fig:supp_corrections}
\end{figure*}

Following the qualitative results in the main paper, Figures~\ref{fig:supp_qualitative} and~\ref{fig:supp_corrections} provide further comparisons on the SemanticKITTI validation set. Figure~\ref{fig:supp_qualitative} shows that RayLift produces more continuous scene layouts and preserves finer scene structures than the compared methods, particularly near road boundaries, roadside objects, and vegetation--terrain transitions. Figure~\ref{fig:supp_corrections} further visualizes the regions corrected by RayLift relative to VoxDet. These corrections are concentrated near observed surfaces and structural boundaries. Green denotes semantic-class corrections in voxels previously predicted as occupied, whereas cyan denotes occupied voxels recovered from previous free-space predictions.

\subsection{Model Limitations}

RayLift currently processes each stereo pair independently and constructs ray evidence from a single frame. It therefore does not exploit temporal continuity or observations accumulated across adjacent frames, which could provide additional geometric evidence for surfaces that are temporarily occluded, weakly textured, or poorly matched by stereo estimation, as illustrated by the failure cases in Figure~\ref{fig:supp_failure}. Future work will explore spatially aligning image features and ray evidence across frames to stabilize candidate surfaces and recover more complete voxel structures. Such an extension must also account for ego-motion and dynamic objects to prevent outdated evidence from being propagated to incorrect spatial locations.

\clearpage

\clearpage

\end{document}